\documentclass[a4paper,10pt,conference,onecolumn]{IEEEtran}
\usepackage[utf8x]{inputenc}
\usepackage{amsfonts}
\usepackage{cite}
\let\labelindent\relax
\usepackage{enumitem}
\usepackage{algorithm,algorithmic}
\usepackage{slashbox}
\usepackage[caption=false, font=footnotesize]{subfig}
\ifCLASSINFOpdf
  \usepackage[pdftex]{graphicx}
  \graphicspath{{img/}}
  \DeclareGraphicsExtensions{.pdf,.jpeg,.png,.jpg}
\else
  \usepackage[dvips]{graphicx}
\fi

\usepackage[cmex10]{amsmath}
\usepackage{stfloats}
\usepackage{tikz}
\usepackage{textcomp}
\usepackage{hyperref}

\newcommand\copyrighttext{%
  \footnotesize \textcopyright 2017 IEEE. Personal use of this material is permitted. Permission from IEEE must be obtained for all other uses, in any current or future media, including reprinting/republishing this material for advertising or promotional purposes, creating new collective works, for resale or redistribution to servers or lists, or reuse of any copyrighted component of this work in other works.}
\newcommand\copyrightnotice{%
\begin{tikzpicture}[remember picture,overlay]
\node[anchor=south,yshift=5pt] at (current page.south) {\fbox{\parbox{\dimexpr\textwidth-\fboxsep-\fboxrule\relax}{\copyrighttext}}};
\end{tikzpicture}%
}

\begin{document}
%

\title{An Empirical Approach for Modeling Fuzzy Geographical Descriptors}

\author{
    \IEEEauthorblockN{
        Alejandro Ramos-Soto\IEEEauthorrefmark{1}, 
        Jose M. Alonso\IEEEauthorrefmark{1},
        Ehud Reiter\IEEEauthorrefmark{2},
        Kees van Deemter\IEEEauthorrefmark{2},
        Albert Gatt\IEEEauthorrefmark{3}}
        \\
    \IEEEauthorblockA{
        \begin{tabular}{ccc}
            \begin{tabular}{@{}c@{}}
                \IEEEauthorrefmark{1}
                    Centro Singular de Investigación\\
                    en Tecnoloxías da Información (CiTIUS),\\
                    Universidade de Santiago de Compostela \\
                    \tt \footnotesize alejandro.ramos@usc.es \\
                    \tt \footnotesize josemaria.alonso.moral@usc.es
            \end{tabular} & \begin{tabular}{@{}c@{}}
                \IEEEauthorrefmark{3}
                    Department of Computing Science,\\
                    University of Aberdeen\\
                    \tt \footnotesize e.reiter@abdn.ac.uk \\
                    \tt \footnotesize k.vdeemter@abdn.ac.uk
            \end{tabular} & \begin{tabular}{@{}c@{}}
                \IEEEauthorrefmark{2}
                    Institute of Linguistics,\\
                    University of Malta\\
                    \tt \footnotesize albert.gatt@um.edu.mt
            \end{tabular}
        \end{tabular}
    }
}

\maketitle
\copyrightnotice

\begin{abstract}
We present a novel heuristic approach that defines fuzzy geographical descriptors using data gathered from a survey with human subjects. The participants were asked to provide graphical interpretations of the descriptors `north' and `south' for the Galician region (Spain). Based on these interpretations, our approach builds fuzzy descriptors that are able to compute membership degrees for geographical locations. We evaluated our approach in terms of efficiency and precision. The fuzzy descriptors are meant to be used as the cornerstones of a geographical referring expression generation algorithm that is able to linguistically characterize geographical locations and regions. This work is also part of a general research effort that intends to establish a methodology which reunites the empirical studies traditionally practiced in data-to-text and the use of fuzzy sets to model imprecision and vagueness in words and expressions for text generation purposes.
\end{abstract}

\IEEEpeerreviewmaketitle

\section{Introduction}
Systems that generate texts from non-linguistic data are known as data-to-text (D2T) systems. One of the main concerns in D2T is to achieve proper definitions of the words and terms which are included in the automatically generated texts. This task is often performed in D2T using analytical studies on sets of human-produced corpus texts and associated data, or through experiments that allow to define the underlying semantics of the generated expressions. For instance, \cite{bib_mousam} provided a thorough analysis of the use of temporal expressions such as ``by the evening'' or ``by midday'' by different forecasters and described their inconsistencies. This analysis led to crisp interval definitions of these expressions, which were subsequently used in the \textsc{SumTime-Mousam} system.

D2T systems such as the aforementioned \textsc{SumTime-Mousam} or RoadSafe \cite{Turner2008, nlg_roadsafe2} generate texts that include vague terms and expressions for referring to time intervals and geographical zones. These expressions are vague in the sense that they do not indicate specific times or zones, e.g, as in ``by the evening'', in opposition to ``between 6pm and 1am''. However, their underlying semantics are defined by means of crisp approaches, such as the time intervals in \textsc{SumTime-Mousam} or the grid-based partition of the geography in RoadSafe.

Research and techniques on fuzzy linguistic descriptions of data \cite{bib_Yager,bib_kacprzyk0,bib_rev_lsummarization} have been proposed as a means to address the problem of extracting linguistic information from data using vague terms inherent to human language \cite{bib_kacprzyk3} for D2T. In this context, there is currently a high interest within D2T for exploring the use of fuzzy sets to, when feasible, better grasp the semantics of vague or imprecise words and expressions. Some examples of this trend include the application of possibility theory to model and convey temporal expressions using modal verbs \cite{bib_gatt_portet_uncertain}, or the use of fuzzy properties in the problem of referring expression generation \cite{bib_dani}. The textual weather forecast generator GALiWeather \cite{bib_galiweather} also used fuzzy sets and fuzzy quantified statements \cite{bib_Zadeh83} to model and compute expressions describing the cloud coverage variable.

Particularly, the work here described will focus on the problem of generating geographical referring expressions \cite{bib_rodrigo}, such as ``Northern Scotland'', ``South of Spain'' or ``Coast of Galicia''. Our setting and departing point is the methodology proposed in \cite{bib_fuzzygre}, that suggests merging the traditional empirical approaches utilized in D2T with the imprecision management capabilities of fuzzy sets and their application in linguistic descriptions of data \cite{bib_role_ldd_nlg,bib_ldd_time_series}.

\begin{figure}[h]
\centering
  \includegraphics[width=0.45\columnwidth]{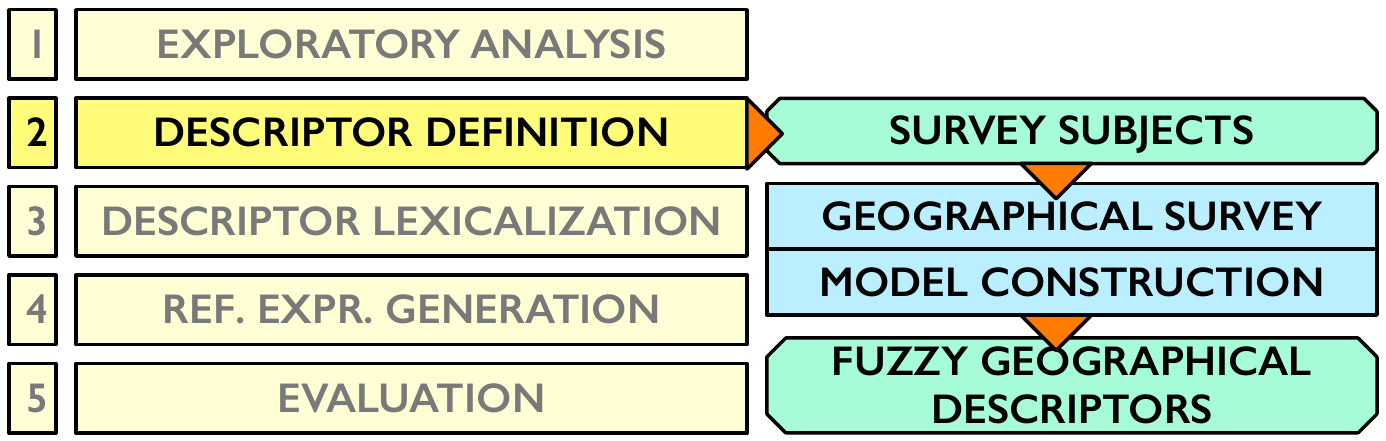}
 \caption{Tasks of the methodology proposed in \cite{bib_fuzzygre}.}
 \label{methodology}
\end{figure}

The methodology described in \cite{bib_fuzzygre} (depicted in Fig. \ref{methodology}), considers a series of tasks that should be performed to generate geographical referring expressions based on fuzzy properties:
\begin{enumerate}
\item An exploratory analysis of the problem from a general perspective (already described in \cite{bib_fuzzygre}).
\item A proper empirical definition of the primitive descriptors, based on data gathered from users.
\item An study on how to lexicalize the possible occurrences of the descriptors (e.g. by means of combination, `north' and `east' = `northeast') and how to generate the referring expressions.
\item An algorithm that implements the referring expression generation strategies determined in the previous step.
\item An evaluation of the algorithm that generates the geographical referring expressions.
\end{enumerate}

This paper describes our approach to the second task, i.e., defining fuzzy geographical descriptors empirically (highlighted in Fig. \ref{methodology}), which can then be used to characterize both individual geographical locations and regions. For this, the rest of the paper is structured in three sections. Section \ref{modeling} encompasses the description of the whole approach, which is tested on a realistic use case. Particularly, in Section \ref{survey} we describe the survey that allowed us to gather data to apply the method described and evaluated in Section \ref{method}. Section \ref{discussion} provides a discussion on several aspects of our approach. Finally, Section \ref{conclusion} provides some ending remarks about this work and points at potential future extensions.

\section{Modeling Fuzzy Geographical Descriptors} \label{modeling}
Our approach for this task is composed of two main elements: a survey that gathers geographical data about the descriptors from users (details are given in Sec. \ref{survey}) and a method that builds their corresponding fuzzy definitions (the methodology is described in Sec. \ref{method}). As in any D2T system, the procedure we have followed corresponds to a knowledge acquisition task. This task is meant to capture the vagueness that, for given linguistic terms (geographical descriptors in our case), arises from having different interpretations from different users.

Although the methodology and our specific approach for modeling fuzzy geographical descriptors are meant to be general and domain-independent, we are following a realistic use case to support this research. Specifically, our use case is focused around some of the most common geographical descriptors which, for instance, have been used in the RoadSafe system \cite{Turner2008,nlg_roadsafe2}. Such descriptors are modeled here in the context of the Galician region (Spain), based on the interpretation of subjects who are assumed to have a minimum knowledge about the Galician geography and that are potential end users (hearers/readers) of automatically generated texts which include geographical references, such as weather forecasts.

\subsection{Description of the survey} \label{survey}
The survey was designed having in mind the lessons learnt from the preliminary experiment described in \cite{bib_fuzzygre}. In consequence, the survey was prepared to be very intuitive and short, without requiring excessive effort from the participants. In a within-subjects design, participants were given 2 different geographical descriptors, and were asked to draw polygons in a map that represented their own interpretation of each descriptor. 

\subsubsection{Materials}
A single map was prepared for the Galician region in Spain, including only the geopolitical data appearing in the source map, which was provided by Mapbox \cite{mapbox} and further customized to show only the political borders of the region, a few location names and the altitude of the terrain.

We included 2 descriptors in the survey. Specifically, we incorporated into this survey two geographical descriptors that were also considered in \cite{bib_fuzzygre}, namely both north and south cardinal directions. Based on these descriptors, the basic expressions ``Northern Galicia'' and ``Southern Galicia'' were provided to the participants (in Spanish).

\subsubsection{Participants}
These were recruited at the high school I.E.S. A Xunqueira I, located in Pontevedra, one of the four province capitals in Galicia. Specifically, students aged between 15 and 17 years old anonymously answered the survey.

\begin{figure}[h]
\centering
  \includegraphics[width=0.6\columnwidth]{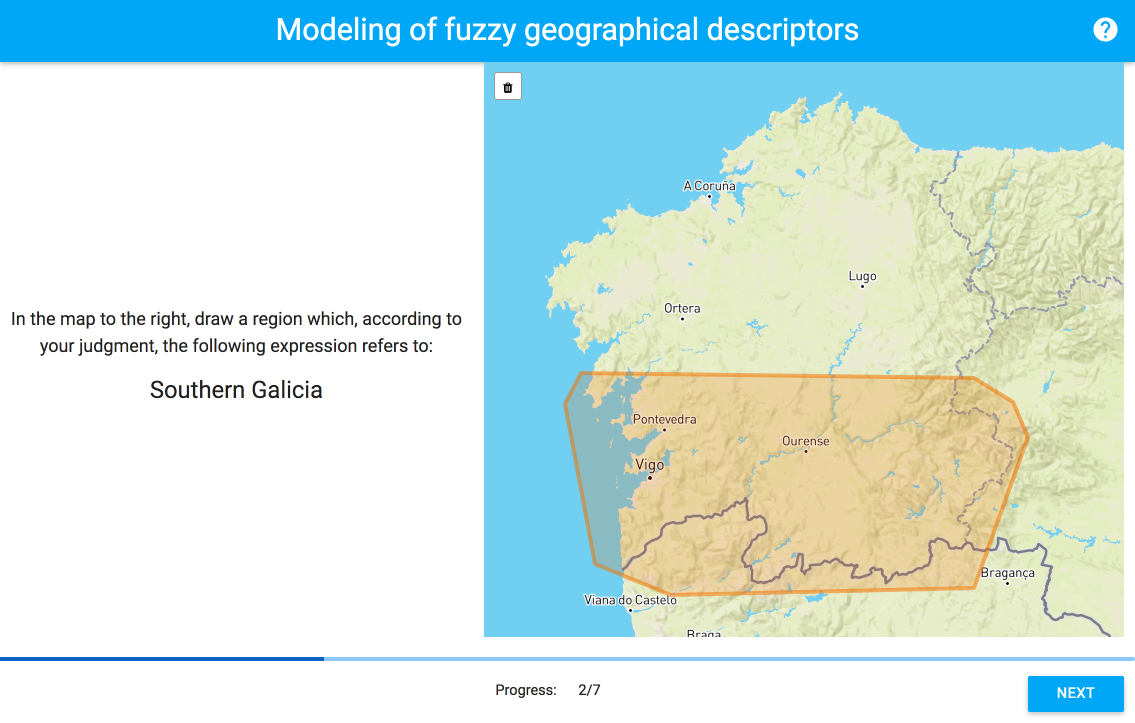}
 \caption{Screenshot of the web interface for the survey (translated from Spanish).}
 \label{itfsurvey}
\end{figure}

\subsubsection{Procedure}
The participants accessed a web interface (see Fig. \ref{itfsurvey}) that provided them with the tools needed to complete the survey. Specifically the participants were asked to draw polygons with limitless points for each geographical expression. The expressions were presented to each participant in random order.

Before starting the survey, the students were required first to draw a simple polygon to get familiarized with the drawing tools. Then, they were allowed to draw only one polygon for each descriptor at a time, although it could be erased and redrawn without restrictions before proceeding to the subsequent descriptor.

After providing responses for all the descriptors, the participants were given the possibility of providing free-text comments about any aspect of the survey.

\subsubsection{Results}
\begin{figure}[h]
\centering
  \includegraphics[width=0.4\columnwidth]{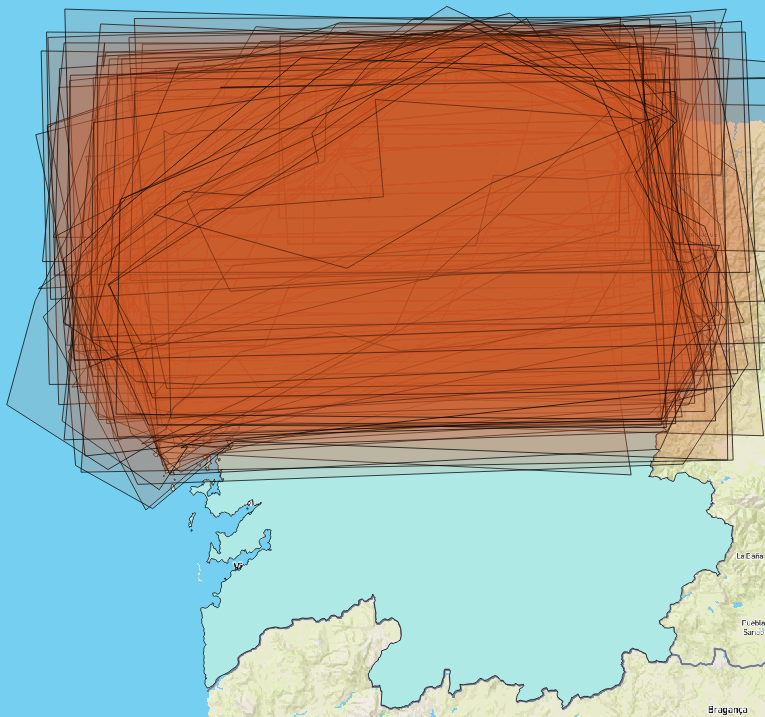}
 \caption{Representation of the responses collected in the survey for the expression ``Northern Galicia''.}
 \label{north}
\end{figure}

We received 99 responses in total for each descriptor, which were inspected visually. For instance, Fig. \ref{north} shows the graphical representation of all the answers for the ``Northern Galicia'' expression, with polygons mainly overlapping towards the upper side of the map and a decreasing density towards the middle of the region.

A pair of the answers provided by the participants were manually removed for different reasons, including a polygon that clearly represented the other descriptor and another with an extremely deformed shape. Thus, we were left with 98 polygons for ``Northern Galicia'' and 98 for ``Southern Galicia''.

\subsection{Construction of Fuzzy Descriptors} \label{method}
The method we have followed to build the fuzzy descriptors is based on the concept of voting model defined by Baldwin in \cite{bib_voting}. Since we are aiming in our case at the modeling of fuzzy geographical descriptors, we required a procedure that is able to consider the bidimensional nature of this domain.

In general terms, our method is based on the construction of a grid of points of a certain granularity, which is then used as the basis of the resulting fuzzy geographical descriptor. The points where most of the polygons drawn by the participants in the survey intersect are given higher membership degrees, while those in the opposite situation get lower membership degrees.

\subsubsection{Method}
The method we propose builds fuzzy geographical descriptors characterized by a membership function that maps a point in the geographical plane (specified in latitude and longitude coordinates) to a specific membership degree. Formally, we model the semantics of a fuzzy geographical descriptor $G$ (e.g., `north'), using a function $\mu_G: \mathbb{R} \times \mathbb{R} \rightarrow [0,1]$.

\begin{algorithm}[h]
 \caption{Fuzzy grid construction for a descriptor $G$}
 \label{fgrid}
 \begin{algorithmic}[1]
 \renewcommand{\algorithmicrequire}{\textbf{Input:}}
 \renewcommand{\algorithmicensure}{\textbf{Output:}}
 \REQUIRE $R_G$, $P_G$
 \ENSURE $MD_G\ |\ \forall\ rp_i \in R_G,\ md_i \geq 0$
 \STATE $MD_G \gets \{\}$
 \FORALL {$rp_i \in R_G$}
    \STATE $count \gets 0$
    \FORALL {$p_j \in P_G$}
        \IF {$(lon_i,lat_i) \subset p_j$}
        \STATE $count \gets count + 1$
        \ENDIF
    \ENDFOR
    \STATE $md_i \gets count$
    \STATE $MD_G \gets MD_G \cup md_i$
  \ENDFOR
  \STATE $max\_md \gets $max$(MD_G)$
  \STATE $\forall\ md_i \in MD_G,\ md_i \gets md_i/max\_md$
  
 \RETURN $MD_G$ 
 \end{algorithmic} 
 \end{algorithm}

In order to compute $\mu_G$, first we define a grid of equidistant points across latitude and longitude, $R_G=\{rp_1,...,rp_i,..., rp_{|R_G|}\}$, which is delimited by the geographical bounds of the underlying geography (the Galician region in our particular case). Each $rp_i$ is a tuple $rp_i = \{lon_i,lat_i\}$, where $lon_i$ and $lat_i$ are the point coordinates ($lon$ stands for longitude and $lat$ for latitude). This grid depends on a granularity parameter that defines the distance between each pair of points in the grid for both latitude and longitude coordinates. Furthermore, the points in the grid are guaranteed to be contained within the shape of the underlying geography.

Using $R_G$ and the list of empirical interpretations defined as polygons for the given descriptor, $P_G = \{p1,...,p_{|P_G|}\}$, the algorithm described in Algorithm \ref{fgrid} determines the membership degrees, $MD_G = \{md_1,...,md_i, md_{|R_G|}\}$ of the grid points in $R_G$. In short, the algorithm counts the number of times that each point is contained in the polygon responses, and then normalizes the result count to provide a real number in the [0,1] interval.

\begin{figure}[h]
\centering
  \includegraphics[width=0.6\textwidth]{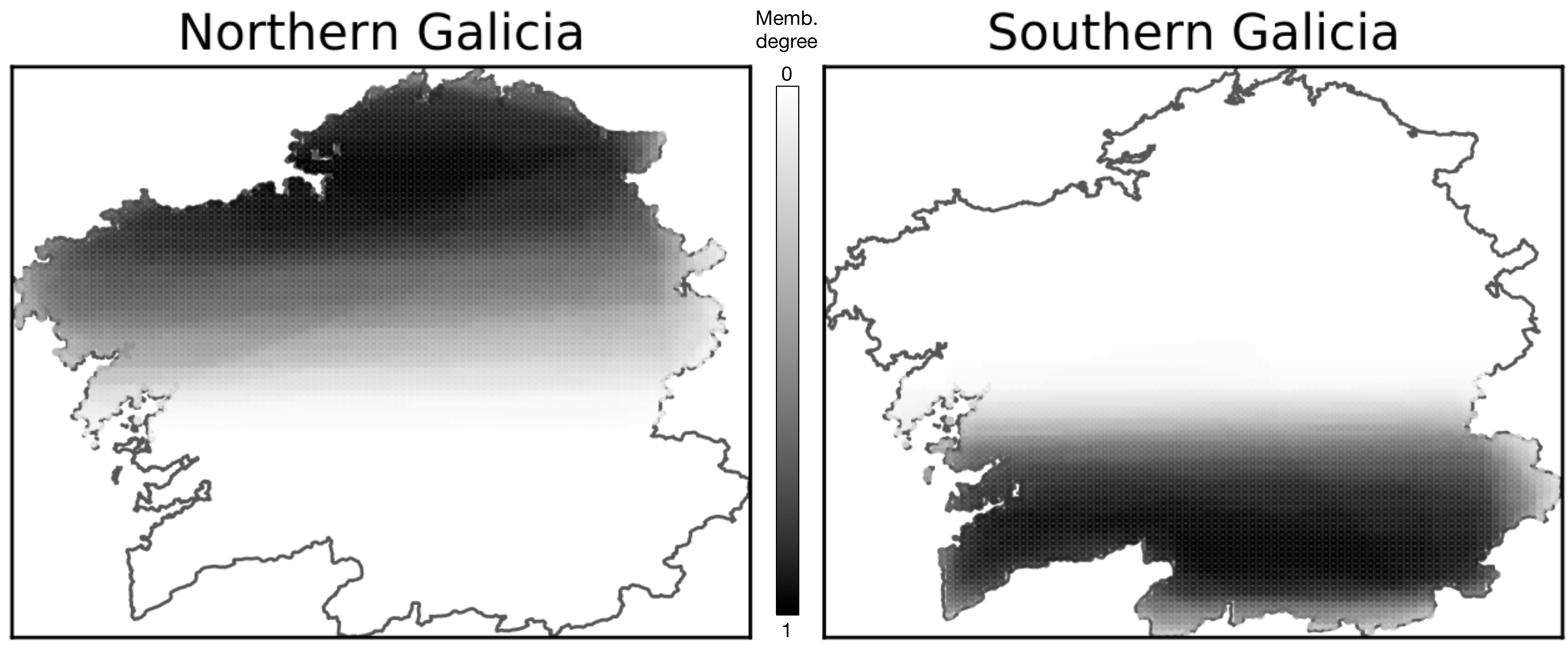}
 \caption{Visual representation of the fuzzy grids obtained using Algorithm \ref{fgrid} from the data gathered in the survey of our use case for ``Northern Galicia'' and ``Southern Galicia''.}
 \label{fgrids}
\end{figure}

Once the membership degrees for $R_G$ have been calculated, it is possible to visualize how the fuzzy grid looks in a map. For instance, for our use case, Fig. \ref{fgrids} shows representations of the fuzzy grids for ``Northern Galicia'' and ``Southern Galicia'' respectively. Specifically, these were built using a granularity distance between points of 1\% in relation to the Galician geographical bounds.

With $MD_G$ already defined, $\mu_G$ adopts the form of a second algorithm that, for any location point $ep$ defined by its coordinates, determines the membership degree, $md_{ep}$ of the location. The procedure, which is shown in Algorithm \ref{feval}, determines the closest 4 points to $ep$ in the grid and calculates the membership degree of the provided location using a weighted interpolation.

This weighted interpolation depends inversely on the distance of the grid points to the evaluated location, and is only performed if the distance from the nearest point in the fuzzy grid to $ep$ exceeds a specific threshold $\epsilon$. Otherwise the membership degree of the nearest grid point is provided.

  \begin{algorithm}[h]
 \caption{Algorithm that evaluates $\mu_G$ for $ep$}
 \label{feval}
 \begin{algorithmic}[1]
 \renewcommand{\algorithmicrequire}{\textbf{Input:}}
 \renewcommand{\algorithmicensure}{\textbf{Output:}}
 \REQUIRE $R_G$, $MD_G$, $ep$, $\epsilon$
 \ENSURE $md_{ep} | 1 \geq md_{ep} \geq 0$
 \STATE $distances \gets \{\}$
 \FORALL {$rp_i \in R_G$, $md_i \in MD_G$}
    \STATE $distances \gets distances \cup \{$distance($rp_i$,$ep$), $rp_i$, $md_i\}$
\ENDFOR
\STATE sort($distances$)
\STATE $distances \gets distances[0:3]$
\IF {$distances[0].distance < \epsilon$}
    \STATE $md_{ep} \gets distances[0].md$
\ELSE
    \STATE $total\_dist \gets$ sum($distances$)
    \STATE $weights \gets \{\}$
    \FORALL {$d \in distances$}
        \STATE $weights \gets weights \cup \ (1-\frac{(total\_dist-d)}{total\_dist})$
    \ENDFOR
        \STATE reverse($weights$)
        \STATE $md_{ep} \gets$ sum($weights.*distances.md$)
\ENDIF
 \RETURN $md_{ep}$ 
 \end{algorithmic} 
 \end{algorithm}

In our case, we have used the Haversine function to calculate the distance between two points in kilometers, as it provides a very reasonable tradeoff between precision and computation complexity. Likewise, we established that $\epsilon = 0.00001$, so that when $ep$ practically shares the same location with a point $rp$ in the fuzzy grid, they also share the same membership degree ($md_{ep} = md_{rp}$).

\subsubsection{Evaluation}
We evaluated the method from two different perspectives. First, we checked the balance between different grid granularities and their computational efficiency by comparing the construction of the fuzzy descriptors using different granularity values. Then, we evaluated how well the fuzzy descriptors would correctly predict the answers provided by the participants in the survey.

In our approach, using lower granularities means building grids that contain more points and thus grasp better the graduality of the descriptors which are built from the human interpretations. However, this also means that the computational time will increase accordingly. As engineering often involves finding a tradeoff equilibrium between precision and complexity, we evaluated if higher granularities could be a reasonable alternative to the 1\% we used as basis.

Specifically, we computed $MD$ for `north' ($MD_{NORTH}$) using a 1\% granularity grid, which is used as our baseline. Then, we recalculated $MD$ at 2, 5, and 10 \% granularities. Table \ref{eval_comp} shows the number of points contained in each fuzzy grid and their associated computation time (we used the Python 3.5 language and the library \textit{shapely} for the geospatial operations in a 2,3 GHz Intel Core i7). Results show that there is an important decrease in points and computation time even from 1\% to 2\%.

\begin{table}[h]
\centering
\caption{Evaluation of the tradeoff between efficiency and approximation level of different granularities.}
\begin{tabular}{| c | c | c | c | c | c |}
\hline
Grid granularity & Points & Time (sec.) & Reduction & Avg. diff. & Std. diff. \\ \hline 
1\% baseline grid & 6372 & 30 & - & - & - \\ \hline 
2\% (int. to 1\%)& 1591 & 7.3 & 4.2x & 0.006 & 0.012\\ \hline
5\% (int. to 1\%)& 250 & 1.2 & 30x & 0.015 & 0.03\\ \hline
10\% (int. to 1\%)& 61 & 0.2 & 150x & 0.03 & 0.04\\ \hline
\end{tabular}
\label{eval_comp}
\end{table}

In order to determine the difference between the 1\% granularity fuzzy grid and the rest of the grids in terms of approximating the interpretations of the survey subjects, we computed $\mu_{NORTH}$ on a 1\% grid using the 2, 5 and 10 \% grids as base. In other words, we created three new 1\% fuzzy grids that, instead of being built directly from the survey data, were interpolated on higher granularity grids using our method. Then, we computed the difference between the base grid and the newly calculated grids. As Table \ref{eval_comp} shows, the average difference of membership degrees between the base grid and the grid built on the 2\% fuzzy grid is 0.006, with a standard deviation of 0.01. This means that it is possible to use a higher granularity grid that is computed 4 times faster while still approximating well the subjects' interpretations.

We performed a second evaluation of our method focused on measuring the performance of the fuzzy descriptors in terms of correctly predicting the answers provided by the survey subjects. This evaluation was made using a 10-fold cross-validation methodology \cite{bib_crossvalidation}. Based on the previous results, the models were built using a 2\% granularity, and we also maintained $\epsilon = 0.00001$. Specifically, 30 randomly located points (taken from a 1\% granularity grid) were selected within each polygon. Then, the points were evaluated through $\mu_{NORTH}$ and $\mu_{SOUTH}$. For each point we selected the descriptor with the highest membership degree and compared it with the descriptor the polygon was meant to model. In case of coincidence we counted a positive hit, and a negative hit otherwise.

Based on this procedure, we calculated the average percentage of positive and negative hits for all answer polygons in a single fold, and then the average of all 10 folds, which are shown in Table \ref{evalres}. We also calculated precision (\% of true positives out of the sum of true and false positives) and recall (\% of true positives out of the sum of true positives and true negatives) values for both descriptors, shown in Table \ref{precall}.

\begin{table}[h]
\centering
\caption{Percentages of descriptor winners against actual references.}
\begin{tabular}{| c | c | c |}
\hline
 & North & South \\ \hline 
\% Hits $\mu_{NORTH}$ & 0.994 & 0.014 \\ \hline 
\% Hits $\mu_{SOUTH}$ & 0.006 & 0.986 \\ \hline
\end{tabular}
\label{evalres}
\end{table}

\begin{table}[h]
\centering
\caption{Precision and Recall values for $\mu_{NORTH}$ and  $\mu_{SOUTH}$.}
\begin{tabular}{| c | c | c |}
\hline
 & Precision & Recall \\ \hline 
 $\mu_{NORTH}$ & 0,987 & 0,502 \\ \hline 
 $\mu_{SOUTH}$ & 0.994 & 0.498 \\ \hline
\end{tabular}
\label{precall}
\end{table}

These results show that the fuzzy descriptors generalize well the interpretations provided in the survey. This is due to several reasons, such as having a high number of answers, but also that both descriptors do not overlap excessively, as the points randomly chosen cover the whole test set polygons, and are not focused on the overlapping part of the descriptors.

In any case, getting a lower percentage of positive hits due to the overlapping between descriptors should not be an issue, especially in the case of terms such as `north' and `south', which are considered natural antonyms (we will comment on this in the next section). It is expected that a set of points included in polygons that were meant to model one descriptor actually have a higher membership degree when evaluated by the other descriptor, as we are not in a crisp context.

\section{Critical Discussion} \label{discussion}
In our use case we performed an elicitation approach of fuzzy geographical descriptors based on a polling method, as reviewed in \cite{bib_measurement}. Given the nature of our problem, we converted a yes/no question into asking the subjects in the survey to determine, for a given descriptor, the region that they clearly consider as part of that descriptor. The fuzziness in the geographical descriptors stems then from the variance between subjects.

The purpose of this specific survey is not to validate, at this stage, any postulate within fuzzy set theory, but rather to seek a proper way to elicite the concept of membership degree for geographical descriptors. Consequently, we are not making assumptions about any axioms or constraints that the fuzzy descriptors should fulfill.

Even so, it is not out of place to remark in advance that properties such as monotonicity can play an important role in ensuring the consistency of the semantics of the descriptors we are defining. In fact, in \cite{bib_mousam} it was evidenced that even inconsistencies may arise with just a few different expert interpretations of the same expression. In order to avoid this problem, the designers of a D2T system may need to post-process the raw empirical definitions and add different constraints in order to make sure that the semantics of the expressions actually make sense.

For instance, in our particular case it seems reasonable that $\mu_{NORTH}$ is monotonic, so that for two points $hl$, located in a higher latitude, and $ll$, in a lower latitude, it holds that $\mu_{NORTH}(hl) \geq \mu_{NORTH}(ll)$. This can also be the case for `south' or for other kind of directional descriptors such as `west' or `east', for which the source of the monotonicity would be different (in a different direction or across the longitude dimension). However, in other cases that do not rely solely on directions, but also on political or cultural knowledge, such as named regions, it could be possible to find interpretations where monotonicity may not apply.

\begin{figure}[h]
\centering
  \includegraphics[width=0.6\textwidth]{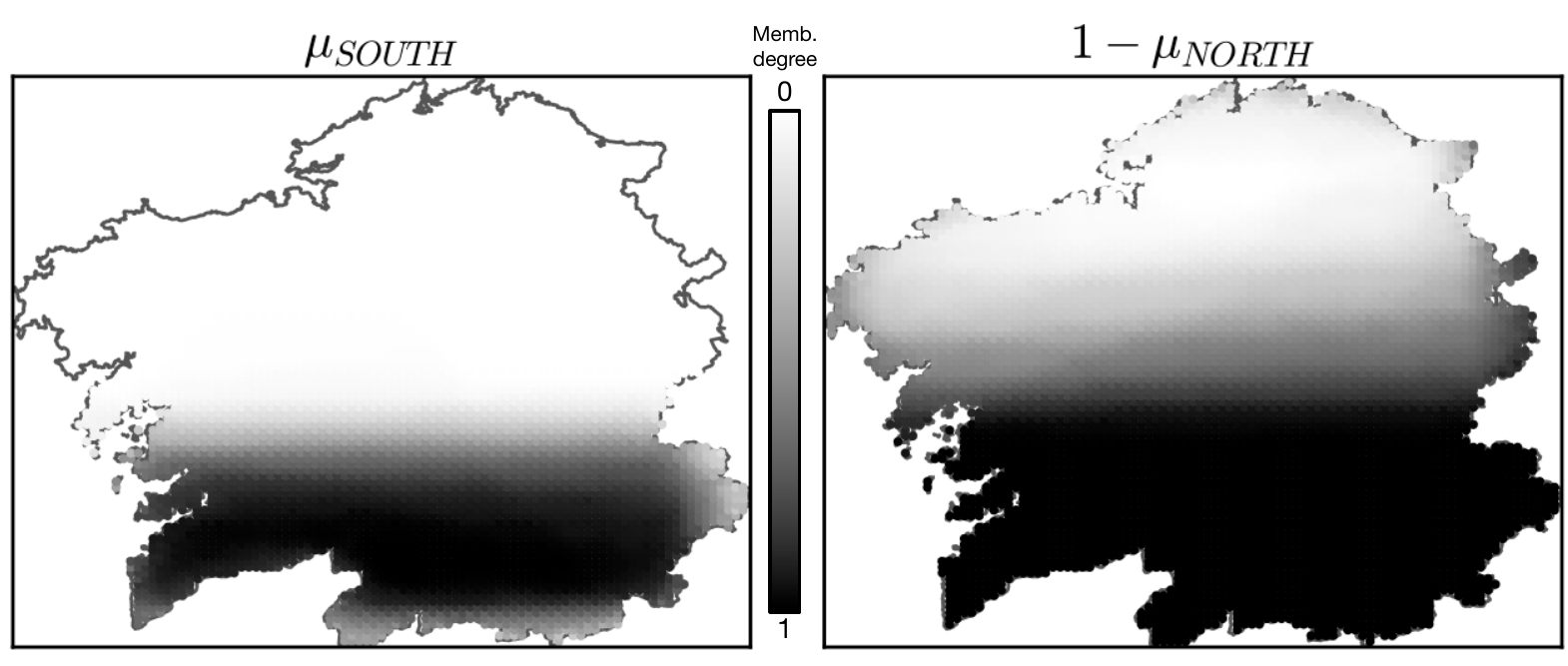}
 \caption{Graphical representation of $\mu_{SOUTH}$ and $1 - \mu_{NORTH}$, interpolated on a 1\% granularity grid using  2\% granularity fuzzy grids as base.}
 \label{antonym}
\end{figure}

Another property which is worth mentioning is the antonymy. For instance, coming back once again to our use case, we consider two geographical descriptors that are in general considered natural antonyms, `north' and `south'. Using the standard negation operator in fuzzy set theory, does it apply then, that $\mu_{SOUTH} \simeq 1 - \mu_{NORTH}$? For illustration purposes, Fig. \ref{antonym} shows the comparison between $\mu_{SOUTH}$ and $1 - \mu_{NORTH}$. Given that we haven't applied any constraints to the fuzzy models built from the survey data, the difference between both descriptors is noticeable visually. In fact, similarities and differences between antonymy and negation in fuzzy logic are an important research topic \cite{bib_antonym_trillas}.

Both monotonicity and antonymy could have been enforced as part of the design of the surveys or experiments that, just like the one described here, allow to gather data about the interpretation of terms and expressions by user or expert subjects. For instance, we could have asked the subjects to draw both `north' and `south' polygons at the same time, or to trace a line that separates both descriptors. However, in our use case we decided to provide more freedom to the subjects to avoid as many biases as possible, even if this meant obtaining less consistent models as a result.

The resulting models tend to be softly gradual, since we were able to gather more than 50 polygons per descriptor. However, the unavailability of subjects or data in general for empirically determining the semantics of words and expressions is a common problem in D2T. This means that in many occasions one will have to do with a few experts or reduced corpora examples. Translating such problem to the context of our approach, it is likely that in real situations we would achieve models of fuzzy geographical descriptors that are closer to being stratified rather than gradual.

On a related note, to decide whether experts or users should provide the interpretation of the terms and expressions to be conveyed in D2T systems is also an interesting problem. This is particularly true in scenarios where experts produce texts or reports whose final audience are not experts themselves, but other groups of users, as in weather forecasts or medical reports for patients.

In such contexts, the meaning of some words and expressions (e.g. ``north'', ``cold'', ``in the morning'') in texts produced by experts does not necessarily match the interpretations of the target audience. This opens the problem of deciding if D2T systems should generate texts that include terms and expressions semantically defined by the experts or by the end users, considering that the latter are meant to understand and make use of the information contained in the texts.

In the case of our survey we did not aim at an expert group in the first place, but focused on a group of potential users of weather forecasts that include the kind of geographical expressions considered in the survey.

\section{Conclusions} \label{conclusion}
We have presented an approach that addresses the problem of defining fuzzy geographical descriptors that aggregate the interpretation of different users, as part of the wider problem of generating geographical referring expressions. For this, we ran a survey that collected data about the geographical interpretation that users make of geographical descriptors such as `north' and `south' in the context of the Galician region in Spain. Based on this dataset, we have proposed a method that builds fuzzy models of the geographical descriptors.

As future work, we intend to run the survey here described to collect the interpretation of expert weather forecasters. This will allow us to obtain deeper insights about the differences between both groups of subjects and study the possibility of merging them into a single model. We will also explore other alternatives for building the fuzzy geographical descriptors, such as the conceptual spaces paradigm \cite{bib_conceptual,bib_dani_color}. Afterwards, we will proceed with the remaining tasks described in the methodology proposed in \cite{bib_fuzzygre}. Specifically, our objective will be to use the primitive fuzzy geographical descriptors to study the generation of actual geographical referring expressions. We plan to incorporate the resulting referring expression generation algorithm into an actual D2T system that produces real-time descriptions of the weather state in the Galician region. In the longer term our aim is to generalize and fully establish this methodology as a standard guideline for the application of fuzzy sets in D2T contexts.

\section*{Acknowledgments}
This work has been funded by TIN2014-56633-C3-1-R and TIN2014-56633-C3-3-R projects from the Spanish ``Ministerio de Economía y Competitividad'' and by the ``Consellería de Cultura, Educación e Ordenación Universitaria'' (accreditation 2016-2019, ED431G/08) and the European Regional Development Fund (ERDF). The authors would also like to thank José M. Ramos González for setting up the survey at the I.E.S. A Xunqueira I, as well as all anonymous subjects who contributed to this study.

\bibliographystyle{IEEEtran}

\end{document}